# Head Frontal-View Identification Using Extended LLE

Chao Wang

Center for Spoken Language Understanding, Oregon Health and Science University

## Abstract

*Automatic head frontal-view identification is challenging due to appearance variations caused by pose changes, especially without any training samples. In this paper, we present an unsupervised algorithm for identifying frontal view among multiple facial images under various yaw poses (derived from the same person). Our approach is based on Locally Linear Embedding (LLE), with the assumption that with yaw pose being the only variable, the facial images should lie in a smooth and low dimensional manifold. We horizontally flip the facial images and present two K-nearest neighbor protocols for the original images and the flipped images, respectively. In the proposed extended LLE, for any facial image (original or flipped one), we search (1) the Ko nearest neighbors among the original facial images and (2) the Kf nearest neighbors among the flipped facial images to construct the same neighborhood graph. The extended LLE eliminates the differences (because of background, face position and scale in the whole image and some asymmetry of left-right face) between the original facial image and the flipped facial image at the same yaw pose so that the flipped facial images can be used effectively. Our approach does not need any training samples as prior information. The experimental results show that the frontal view of head can be identified reliably around the lowest point of the pose manifold for multiple facial images, especially the cropped facial images (little background and centered face).*

## 1. Introduction

Most literatures [1-2] indicate that frontal-face based recognition performs better than those based on non-frontal faces. Therefore, frontal faces from facial video sequence should be identified before subsequent recognition. Recently, some researchers [3-6] exploited the pose manifold, represented as the underlying geometry structure information of the pose data space, to estimate the head poses including frontal view. BenAbdelkader [3] proposed a taxonomy of methods that correspond to different ways of incorporating the pose angle information at the different stages of supervised manifold learning, such as neighborhood graph construction. Ptucha and A. Savakis [4] used facial feature points rather than the whole face image to construct the manifold for pose estimation. Hu et al. [5] noticed that the data points related to different yaw poses of the same person can construct an ellipse-like circle, which differs in rotation degree, size and center position for different persons. Balasubramanian and Ye [6] presented a biased manifold embedding framework in which the head pose information is used to compute a biased neighborhood of each training sample in the feature space before manifold learning. However, there are three limitations for recent manifold based head pose estimation algorithms. (1) They are all supervised methods, which need a large number of face images under different known poses as training samples to construct a precise pose parameter map. These training samples are often difficult or expensive to collect. (2) Due to the difference of face identity between training samples and testing samples, these algorithms require various preprocessing methods to eliminate the identity information of face image before manifold learning. (3) In order to precisely estimate the head pose, some algorithms use the manually cropped facial images (with very little background and centered face) [3-5].

In this paper, we present an extended Locally Linear Embedding (LLE) [7] for head frontal-view identification in an unsupervised fashion. Many studies [5, 6, 8] indicate that when the pose distributions of the facial images are symmetric (e.g., yaw pose angles varying from -90° to +90° in increments of 2°), the gradient of the frontal-view position in the geometry of pose manifold should be zero (e.g., lowest point of the pose manifold [8]). In order to construct the symmetric distribution of pose, we horizontally flip the facial images. However, due to the background difference, different face position and scale in the whole image and some asymmetry of left-right face, for the same yaw pose, there is still a big distance between the original facial image and the flipped facial image (e.g., 60° yaw pose original facial image vs. 60° yaw pose flipped face image derived from -60° yaw pose original facial image). In order to address this problem, we present two K-nearest neighbor protocols for the original facial images and the flipped facial images, respectively. For any facial image (original or flipped), we search (1) Ko nearest neighbors among the original facial images and (2) Kf nearest neighbors among the flipped facial images to construct the neighborhood graph of the extended LLE. Different reference facial image corresponds to different variable Ko and Kf, which depends on the pose distributions of the original facial images.

## 2. The original LLE algorithm

As a dimension reduction algorithm, the original LLE constructs a neighborhood preserving mapping while couples across all data points. Therefore, LLE uses overlapping local information to discover global structure.

There are three steps of the original LLE [7], described as follows

Step 1. *Construct Neighborhood Graph G: For each data point $\vec{x}_i$, if point $\vec{x}_j$ is one of the K nearest neighbors measured as the Euclidean distance $d_E(\vec{x}_i, \vec{x}_j)$, then connect them with $w_{ij}$ as the edge weight.*

Step 2. *Compute Weight Matrix $W = \{w_{ij}\}$: For each data point $\vec{x}_i$, minimize the cost function*

$$\varepsilon(W) = \sum_i \left| \vec{x}_i - \sum_j w_{ij} \vec{x}_j \right|^2 \quad (1)$$

*by constrained linear fits.*

Step 3. *Construct Low Dimensional Embedding: Compute the vectors $\vec{y}_i$ best reconstructed by the weights $w_{ij}$, minimizing the embedding cost*

$$\emptyset = \sum_i \left| \vec{y}_i - \sum_j w_{ij} \vec{y}_j \right|^2 \quad (2)$$

*by its bottom nonzero eigenvectors.*

## 3. Extended LLE for head frontal-view identification

In this section, we first describe why and how to use LLE algorithm for head frontal-view identification. Then, we present the extended LLE with two K-nearest neighbor protocols for original and flipped image sets, respectively. Lastly, we illustrate the algorithm steps of extended LLE. In this paper, we only consider the yaw pose variation.

### 3.1. Characteristics of manifold geometry using LLE

As described in some recent studies [6, 8], when the pose distribution of facial images is symmetric, the position corresponding to frontal view is located in the center of the manifold geometry using LLE. Due to the yaw pose being the single variable, the manifold geometry in a 2D embedding space should be smooth and one-dimensional. As shown in Figure 1 (a), for a symmetric pose distribution of facial images, the one-dimensional manifold geometry is like a parabola shape and the position corresponding to frontal view is in the vertex. Therefore, we can easily identify the frontal view with the manifold geometry in this situation. When the pose distribution of facial images is asymmetric, the one-dimensional manifold is also like a parabola shape but the position corresponding to frontal view deviates from the vertex (see Figure 1 (b)), which is difficult for head frontal-view identification. From the local view, each data point only covers the K nearest neighbors; however, from the global view, all of the data points are coupled across in the manifold learning procedure, which is influenced by the global distribution of the poses. Isomap [9] has the similar characteristics [5, 6], but the manifold geometry using Isomap is not smoother than that using LLE.

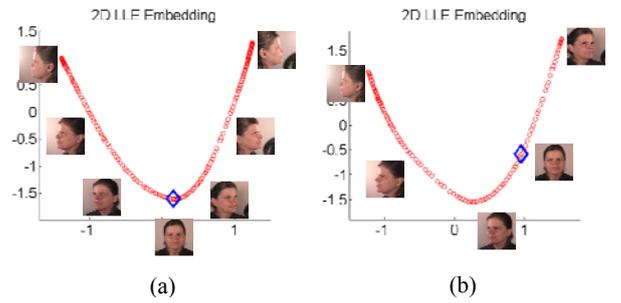

(a) (b)

Figure1: 2D LLE embedding of (a) the facial images with yaw poses varying from -90º to 90º in increments of 1º and (b) the facial images with yaw pose varying from -90º to 30º in increments of 1º ("◊" represents the position corresponding to frontal-view in manifold.)

Generally, the facial images are asymmetric in pose distribution. In order to identify the head frontal view, we can first convert the asymmetric distribution into a symmetric distribution. The simplest way is to horizontally flip the original facial images. In Figure 2, for example, we take some original and corresponding flipped facial images

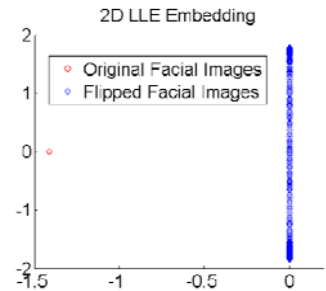

Figure2: 2D LLE embedding of original and flipped facial images to learn manifold using LLE.

We can see that there is a longer distance between the original facial images and the flipped facial images than the distance among themselves. This is because there are some differences, such as different background, various face position and scale in the whole image and some asymmetry of left-right face, between the original facial image and the flipped facial image at the same yaw pose, which are larger than the pose difference among the K nearest neighbors in the original facial images or the flipped facial images.

### 3.2. Two K-nearest neighbor protocols

In order to eliminate the unnecessary differences (noise) between the original images and the flipped facial images, we present two K-nearest neighbor protocols for the original or flipped set of the facial images, respectively. For any data point (facial image), we search Ko nearest neighbors in original set and Kf nearest neighbors in flipped set. In this paper, the total number of Ko and Kf equals to Kt, which is a constant.

However, the variables Ko and Kf can vary according to different reference facial image (data point). For example, there are some original facial images with yaw pose varying from -90° to 30° in increments of 20°; therefore, the yaw poses of the corresponding flipped facial images vary from -30° to 90° in increments of 20°. As shown in Figure 3, we illustrate the novel neighborhood graph based on the image distance (e.g., the image distance between the flipped facial image with 30° yaw pose and the original facial image with 10° yaw pose is 22). If we search the Ko and Ko nearest neighbors of the original facial image with -90° yaw pose, we find that all of the flipped facial images are far away in image distance. Therefore, for this data point, the variable Kf should be less than Ko (some neighbors in flipped set are abandoned; the same number of neighbors in original set are reconfirmed, which keep Kt constant).

Lastly, we describe the rule of abandoning and reconfirming the neighbors in the neighborhood graph. For different data points in the same set (original or flipped), if the same neighbors are searched in the other filed, we only keep the neighbors with the shortest distance (related to the same data point). For those data points which have abandoned neighbors, we need reconfirm the same number of neighbors in the set of the data points. As shown in Figure 3, for the data points with yaw pose -90°, -70°, -50° and -30° in the original set, they have the same neighbors (-30° and -10°) in the flipped set. We only keep the neighbors related to the data point with yaw pose -30° (23+22 is the shortest distance among them). In order to Keep Kt (Ko + Kf) constant, we add new neighbors (-30° and -10°) related to the other data points in the original set.

### 3.3. The extended LLE

Incorporating the two K-nearest neighbor protocols into LLE, we propose an extended LLE for head frontal-view identification. We convert the asymmetric distribution of the original facial images into a symmetric distribution by horizontally flipping the original facial images. Then, we use the two K-nearest neighbor protocols to eliminate the unnecessary differences between the original facial images and the flipped facial images. Besides, the proposed algorithm does not need any training samples.

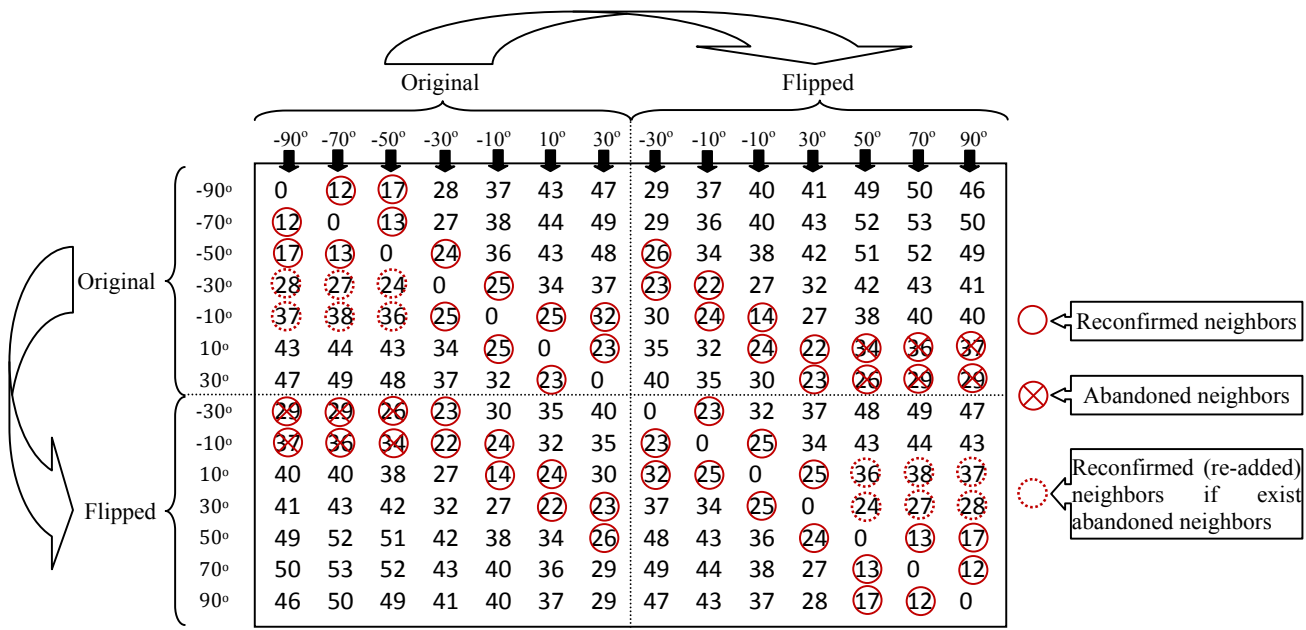

Figure 3: Example of the proposed neighborhood graph. The original facial images with yaw pose varying from -90° to 30° in increments of 20°; the corresponding flipped facial images with yaw pose varying from -30° to 90° in increments of 20°. For any facial image, search the Ko (initially, Ko=2) and Kf (initially, Kf=2) nearest neighbors in original and flipped set, respectively (in the corresponding column). If exist reconfirmed neighbors, the variable Ko and Kf should vary.

Similar to original LLE [7], there are also three steps of the proposed extended LLE, described as follows

Step 1. *Construct Neighborhood Graph G:*

(i). *Flip the original facial images horizontally.*

(ii). *For each data point $\vec{x}_i$ in the original and flipped facial images, perform the two K-nearest neighbor protocols, if point $\vec{x}_j$ is one of searched neighbors measured as the Euclidean distance $d_E(\vec{x}_i, \vec{x}_j)$, then connect them with $w_{ij}$ as the edge weight.*

Step 2. *Compute Weight Matrix $W = \{w_{ij}\}$: For each data point $\vec{x}_i$, minimize the cost function Eq. (1) by constrained linear fits.*

Step 3. *Construct Low Dimensional Embedding: Compute the vectors $\vec{y}_i$ best reconstructed by the weights $w_{ij}$, minimizing the embedding cost Eq. (2) by its bottom nonzero eigenvectors.*

Different from original LLE, the proposed extended LLE considers not only the original facial images but also the flipped facial images. We search K neighbors in the original set and the flipped set, respectively. However, they construct the same neighborhood graph of the extended LLE.

## 4. Experiments

In this paper, we compared the original LLE and the extended LLE on the some facial images with the asymmetric distribution of yaw pose for head frontal-view identification. Besides, we also perform the extended LLE on the corresponding cropped facial images (with little background and centered face).

### 4.1. Facial database

We used FacePix face database [10] to form various videos for experiments. FacePix face database contains 30 individuals with yaw pose angles varying from -90º to 90º in increments of 1º. So for each individual, there are 181 face images, and some examples are shown in Figure 4 (a). The resolution of the face images are 128×128, and the grayscale pixel intensity feature space of the face images are used for experiments. For example, we can construct an image sequence in which head rotates from -90º to 38º in increments of 2º, described as -90º:2º:38º. We also make some cropped facial images based on FacePix face database (see Figure 4 (b)).

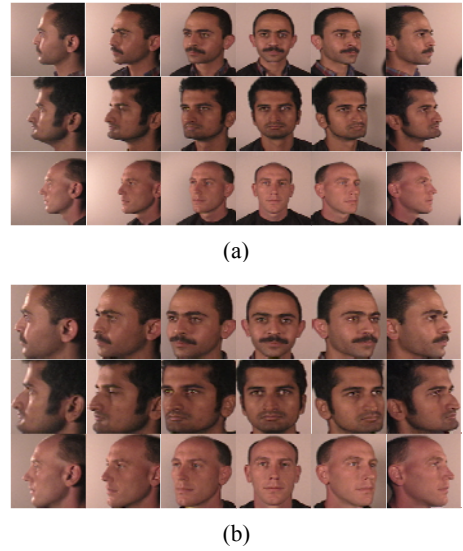

(a)

(b)

Figure 4: Examples of three persons in (a) FacePix face database and (b) cropped FacePix face database. The angles of yaw pose from left to right column are: -90º, -60º, -30º, 0º, 30º and 60º, respectively. Different rows correspond to different persons.

### 4.2. Comparisons and Results

Based on two image sequences in asymmetric pose distribution (-90º:1º:60º and -90º:1º:30º), we compare (1) Case I: the original LLE on the original images, (2) Case II: the extended LLE on the original images and the corresponding flipped images and (3) Case III: the extended LLE on the cropped original and flipped images. For head frontal-view identification, the actual degree of frontal view (0º in each video sequence) is denoted as $v$; the identified degree of frontal view (the lowest point of manifold geometry) is denoted as $w$. The expression $|w - v|$ can represent the identification error. Therefore, we can obtain the identification accuracy (mean $u$ and standard deviation $\sigma$) of head frontal view on FacePix face database (30 individuals), listed in Table 1. Then, $u = E(|w - v|)$ where $E(\cdot)$ represents expectation and $\sigma = \sqrt{E[(|w - v| - u)^2]}$.

Table 1
Estimating accuracy ($u \pm \sigma$) for head frontal view on FacePix face database (30 individuals)

|  | **Case I** | **Case II** | **Case III** |
| --- | --- | --- | --- |
| **-90º:1º:60º** | 10.1º±5.7º | 5.8º±3.6º | 4.2º±3.1º |
| **-90º:1º:30º** | 22.8±5.1º | 5.2º±2.9º | 4.5º±3.4º |

We can see that the extended LLE has high effectiveness for head frontal-view identification (Case II and Case III), no matter how asymmetric the pose distribution is. In addition, for the cropped facial images (Case III), the identifying accuracy using the extended LLE is improved further, although the improvement is not very obvious. This illustrates that even though the facial images are not cropped, the extended LLE can also have a high performance. As shown in Figure 5, we list the experimental results of three persons (in Figure 4) in the comparison.

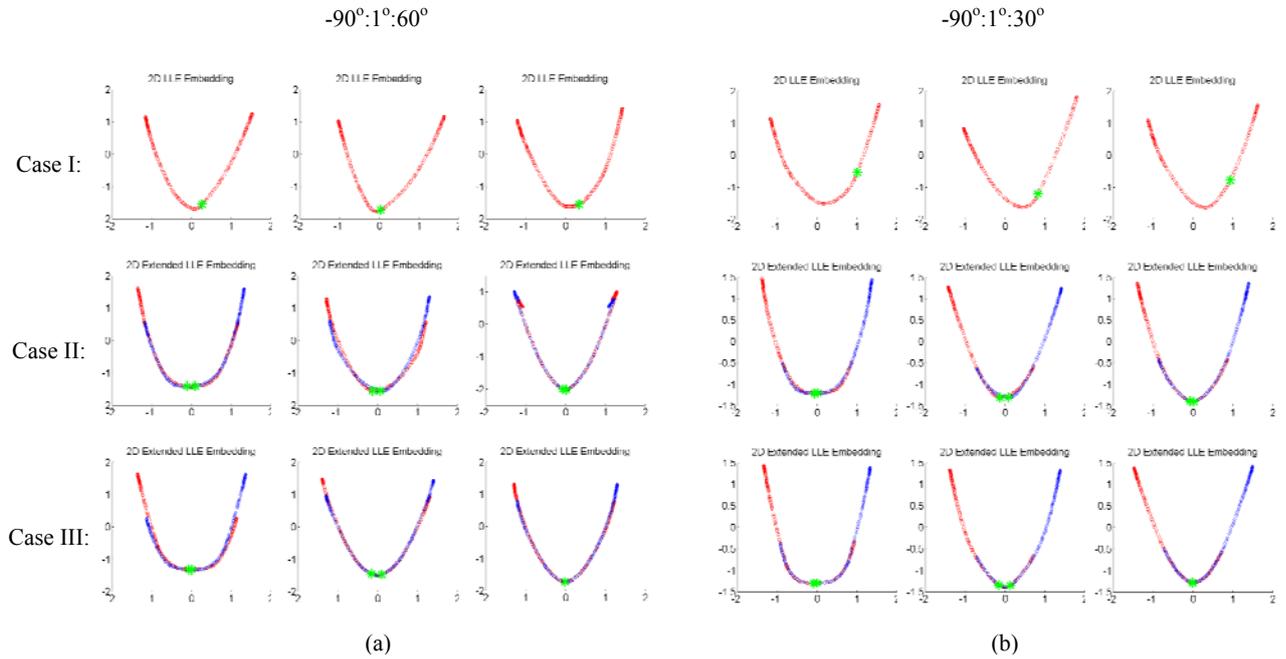

Figure 5: Examples of the comparisons on (a) -90°:1°:60° and (b) -90°:1°:30°. Different rows correspond to different cases. In (a) or (b), the same column corresponds to the same person. (The mark "*" represents the position corresponding to frontal view in manifold.)

## 5. Conclusion and future work

We proposed an extended LLE for head frontal-view identification in an unsupervised fashion. For the facial images with the symmetric distribution of yaw poses, the position corresponding to frontal view can be identified from the manifold geometry using LLE. Based on the characteristic property, we can convert any asymmetric pose distribution into a symmetric pose distribution by horizontally flipping the original facial images. However, there are some unnecessary differences (different background, varying facial position and scale in the whole image and some asymmetry of left-right face) between the original facial images and the flipped facial images, which influence the precision of head frontal-view identification. To address this problem, we extend LLE with two K-nearest neighbor protocols: search the K nearest neighbors in the original facial images and the flipped facial images, respectively. Besides, the proposed algorithm does not need any training samples for head fontal-view identification.

However, there is still room for improvement to do for the extended LLE. It takes much time to manually crop the facial images. We can use the facial appearance automatically obtained by View-based Active Shape Models [11] or Active Appearance Models [2].